%% file: samplepaper.tex
\documentclass[runningheads]{llncs}
\usepackage[T1]{fontenc}
\usepackage{graphicx}
\usepackage{booktabs} 
\usepackage{subcaption}
\usepackage{threeparttable}
\usepackage{amsmath}
\usepackage{amsfonts}
\usepackage{comment}

\begin{document}
\title{Constrained Hyperparameter Optimization for Streaming Data}
%
%
\author{Bruno Veloso\inst{1,2}\orcidID{0000-0001-7980-0972} \and
João Gama\inst{1,2}\orcidID{0000-0003-3357-1195}} 
%
\authorrunning{Veloso and Gama}
%
\institute{FEP - School of Economics and Management, University of Porto , Porto, Portugal \and
INESC TEC, Porto, Portugal
\email{\{bveloso,jgama\}@fep.up.pt}\\
}
\maketitle              
\begin{abstract}
Optimization of hyperparameters is a critical factor to obtain optimal model performance. While existing research has predominantly concentrated on batch-learning scenarios, addressing the complexities inherent in data streams presents a challenge.
The deployment of sophisticated methodologies to manage data streams becomes highly important. Consequently, the capacity for self-adjusting hyperparameters during online learning phases emerges as a goal.

Many hyperparameters exhibit constraints and are confined within bounded search spaces, rendering specific solutions unacceptable upon applying optimization operators. To solve this issue, employing boundary constraint-handling techniques becomes imperative to rectify invalid solutions. This paper presents strategies for effectively managing boundary constraints within constrained numerical optimization problems. Recent methodologies, including heuristic and evolutionary-based optimization, employ a "boundary" strategy, wherein values that surpass boundary thresholds for a given hyperparameter are realigned to the respective limits.

Our study introduces four strategies to navigate boundary constraints in online optimization algorithms. Through empirical investigations conducted on established datasets, we demonstrate that adopting boundary strategies outperforms the "boundary" strategy.
\end{abstract}

\input{samplebody-conf}
%
%
%
\bibliographystyle{splncs04}
\bibliography{sample-bibliography} 
%

\end{document}

%% file: samplebody-conf.tex
\section{Introduction}
\label{sec1}

The rapid advancement of communication technology has led to an increase in data generation, opening up doors to various applications, including predictive maintenance, healthcare monitoring, and automation. Extracting valuable insights from these dynamic data streams is crucial for the private and public sectors. At the same time, automated machine learning (AutoML) has seen significant growth in the past two decades, focusing on hyper-parameter optimization, resulting in various algorithms, such as grid search \cite{lerman1980fitting}, random search \cite{bergstra2012random}, Bayesian Optimisation \cite{mockus1978application}, Evolutionary Algorithms \cite{back1996evolutionary}, and Swarm Algorithms \cite{kennedy1995}, to optimize hyper-parameters.

Most offline process automation methods rely on training with static data batches. However, these supervised models often need help with concept drift. Adapting the model to the current data distribution requires restarting the tuning process whenever a concept drift occurs \cite{gama2014}. Hence, progress in online AutoML, especially in hyper-parameter self-tuning, is crucial. While research in this area is limited, some notable approaches address fundamental aspects and challenges of evolving data streams \cite{Bahri2021}.
Some approaches adapt traditional offline learning to an incremental process \cite{Bakhashwain2020212}, while others suggest restarting hyper-parameter tuning after a concept drift \cite{Veloso0MV21,moya2023improving}. 

The optimization process involves systematically exploring the search space to identify optimal values of the objective function. However, such exploration may inadvertently cause solution vectors to violate their specified bounds, yielding invalid results. Numerous methods for handling boundary constraints have been proposed in the literature to address this issue, particularly in batch learning.

In this study, we have adapted heuristic-based (Nelder-Mead \cite{Veloso0MV21}) and evolutionary-based (micro-evolutionary \cite{moya2024improving}) online optimization algorithms to better accommodate boundary-constrained optimization problems. Our primary objectives include conducting a single pass over the data, effectively detecting and responding to concept drift, and rectifying invalid configurations generated by these algorithms to be coherent with the boundary constraints.

The boundary constraint methods adapted and integrated into the online hyper-parameter optimization algorithms include Centroid \cite{juarez2017improved}, Random \cite{price2006differential}, Reflection \cite{ronkkonen2005real}, and Wrapping \cite{purchla2004experimental} to restore invalid vectors to the admissible region. Despite their efficacy, these methods were not originally devised for online constrained problems, where solutions must remain within the admissible region throughout the optimization process.

The principal contributions of our proposal include:
\begin{itemize}
    \item Utilizing advanced hyper-parameter optimization techniques, two state-of-the-art optimizers were adapted to incorporate boundary constraint strategies. 
    \item Subsequently, comprehensive and rigorous evaluations were conducted, using different datasets and employing diverse boundary constraint methodologies.
\end{itemize}

The document is organized into five sections. Section \ref{sec2} presents the related work, focusing on online hyper-parameter optimization algorithms and boundary constraint strategies. Section \ref{sec3} describes hyper-parameter optimization algorithms and the adopted boundary constraint strategies. Section \ref{sec4} presents the experiments and results, while Section \ref{sec5} consolidates the conclusions.

\section{Related Work}
\label{sec2}
While Auto Machine Learning (AutoML) has been explored for decades for algorithm selection, contributions in the literature related to online hyper-parameter tuning are much more recent. Our literature search was focused on online hyper-parameter optimization and Boundary Constraints.

\subsection{Hyper-parameter tuning for data streams}

In recent research, there has been an increase in interest in online hyper-parameter optimization, particularly within the framework of data streams. \cite{zhan2018efficient} and \cite{Lin20196578} have integrated hyper-gradient-based optimization methodologies into online learning frameworks. However, it is pertinent to highlight that these methodologies, as mentioned by \cite{imbrea2021automated}, do not explicitly accommodate the influence of concept drifts. \cite{imbrea2021automated} further applied AutoML tools to real-time streaming data, highlighting the imperative need to adapt to concept drift events for better data management. \cite{Veloso2018241} introduced the Self Hyper-Parameter Tuning (SPT) technique, employing Nelder-Mead optimization within dynamically adjustable windows for instantaneous configuration recognition. Subsequently, \cite{Veloso0MV21} refined this approach employing a single-pass algorithm in SPT to improve the responsiveness to concept drifts in near-real-time scenarios.

Lacombe et al. \cite{Lacombe2021} proposed a framework that suggests hyper-parameter values based on prior knowledge, optimizing the Adaptive Random Forest and drift detection parameters.

Two studies incorporated evolutionary optimization methodologies for online learning. Bakhashwain et al. \cite{Bakhashwain2020212} employed a genetic algorithm for hyper-parameter optimization in a deep long short-term memory model, focusing on dynamic optimization but omitting explicit treatment of concept drifts. Kulbach et al. \cite{Kulbach2022472} adapted the CASH problem to an online scenario using a genetic algorithm, particularly detecting and adjusting to concept drifts. More recently, \cite{moya2023improving} proposed the MESSPT algorithm for data streams; the proposed method follows the principles of the SPT algorithm advanced by \cite{Veloso0MV21} but substitutes the heuristic-based approach with a micro-evolutionary technique.
Liu et al. \cite{liu2023online} proposed a framework for Class-incremental learning where the number of classes increases during the data stream. The authors formulate the hyper-parameter optimization process as an online Markov Decision Process and apply locally estimated rewards and a bandit algorithm to generate the solutions.

\subsection{Boundary Constrains}

Various methodologies are developed for addressing boundary constraints in optimization problems. Notable among these strategies is the Boundary method, advocated by \cite{brest2006self}, wherein decision variables transgressing delimited bounds are either reset to the violated bound min or max values. The Reflection technique, as proposed by \cite{robinson2004particle}, involves mirroring the values back from the violated bound, while the Wrapping approach, described by \cite{zhang2004handling}, comprises reflection from the opposite violated bound. Additional methodologies like the Random method described by \cite{price2006differential} includes perturbing variables within a random range.

However, none of these approaches were explicitly developed to address constrained online optimization problems, characterized by a defined feasible region where the admissible solutions must reside exclusively. A recent advancement in this domain is the introduction of the Centroid method by \cite{juarez2017improved}. This innovative technique involves positioning the corrected vector at the centroid formed by k + 1 solution vectors, one of which is drawn from the population proximate to the admissible region, with the remaining k vectors subjected to random correction procedures.

Ye et al. \cite{ye2022online} proposes the reduction of inter-domain discrepancy between the source data instances and the unlabelled target instances that arrive as a data stream. To reduce the intra-domain discrepancy, the authors used a classifier trained on a set of known instances with respective labels. Then, they applied a boundary constraint to this set of instances to enhance the classifier recognition performance. In our work, the hyper-parameter boundary constraints are already defined by the selection of the model that we want to optimize. 

Balcan et al. \cite{balcan2024new} proposes tuning regularization parameters in regularized logistic regression. The authors propose an upper bound on the approximation error between the original and approximated loss functions to obtain a learning guarantee.


In brief, the optimization methodologies available via online platforms can be distinguished into two primary categories: static and dynamic strategies. The static strategies comprehend periodic or single optimization processes. The dynamic strategies employ drift detection mechanisms to restart the optimization process, but face challenges during the exploration phase due to the inherently dynamic nature of data streams. Additionally, the efficacy of optimization algorithms in scenarios with boundary constraints, wherein hyper-parameters operate within predefined bounds, remains a pertinent concern. Acknowledging this literature gap, our proposal focuses on a comprehensive investigation into the implications of employing boundary constraint strategies on online optimization methodologies.

\section{Proposed Method}
\label{sec3}

\begin{figure*}[!ht]
     \centering
     \begin{subfigure}[t]{0.18\textwidth}
         \centering
         \includegraphics[width=\textwidth]{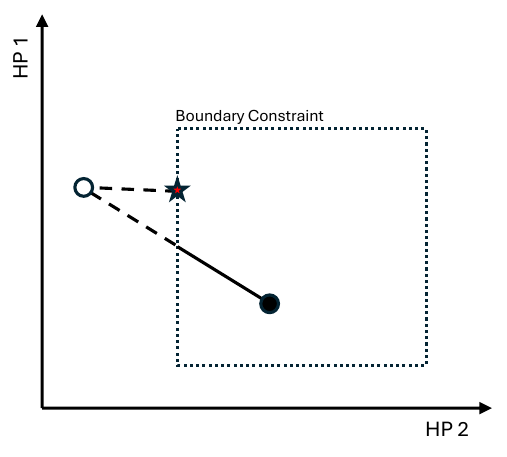}
         \caption{Boundary}
         \label{fig:boundaryx}
     \end{subfigure}
     \begin{subfigure}[t]{0.18\textwidth}
         \centering
         \includegraphics[width=\textwidth]{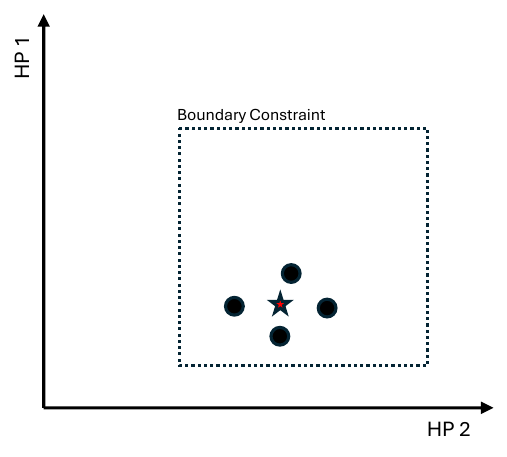}
         \caption{Centroid}
         \label{fig:centroidx}
     \end{subfigure}
     \begin{subfigure}[t]{0.18\textwidth}
         \centering
         \includegraphics[width=\textwidth]{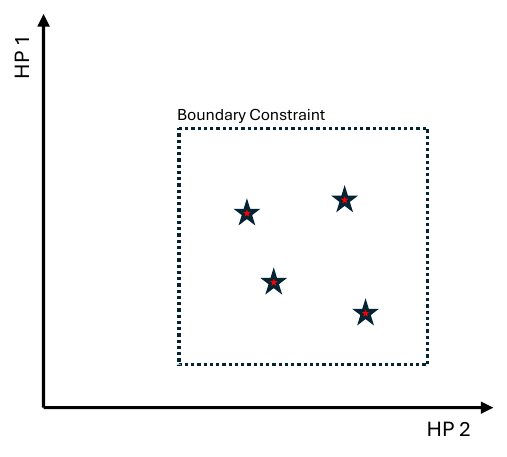}
         \caption{Random}
         \label{fig:randomx}
     \end{subfigure}
        \begin{subfigure}[t]{0.18\textwidth}
         \centering
         \includegraphics[width=\textwidth]{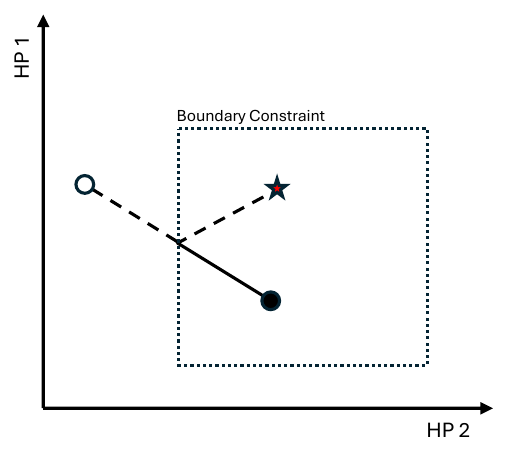}
         \caption{Reflection}
         \label{fig:reflectionx}
     \end{subfigure}
     \begin{subfigure}[t]{0.18\textwidth}
         \centering
         \includegraphics[width=\textwidth]{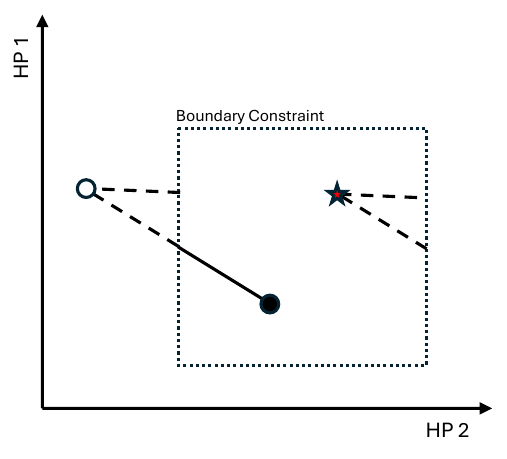}
         \caption{Wrapper}
         \label{fig:wrapperx}
     \end{subfigure}
     \par\bigskip
        \caption{Boundary Constraint Strategies -- Black circles - previous configurations; White circle - Calculated new configuration; Star - Final configuration after applying the strategies}
        \label{fig:strategies}
\end{figure*}

In this section, we will describe two online hyper-parameter tuning methods as well as some boundary constraint techniques. We have designed our approach to align with the less computationally intensive methods documented in literature (boundary, centroid, random, reflection and wrapper presented in Figure \ref{fig:strategies}. For the sake of reproducibility, the corresponding codebase is publicly accessible on GitHub (anonymized).

\subsection{Online Hyper Parameter Tuning}

The problem of hyper-parameter tuning can be formulated by: 
A data stream $s$, a machine learning algorithm $A$ with its hyper-parameters $A_{hp}$ ($hp$ of the hyper-parameters extracted from a set of possible hyper-parameters $H( \_ , \_, \_ )$) and $L$ a loss metric. Considering a batch of n-dimensional elements $x_i \in \mathbb{R}^d $ with $i = 1, \dots, n$ extracted at each step from stream $s$ and the target value related to each element $y_i$, our objective is to find the best $A_{hp}^*$ which satisfies that: $A_{hp}^* = \underset{hp \in H }{argmin} \dfrac{1}{n} \sum_{i=1}^n L(A_{hp}, x_i, y_i)$.


The Self Hyper-Parameter Tuning (SPT) algorithm, as described by \cite{Veloso0MV21}, operates through two distinct modes: exploration and exploitation. During the exploration phase, the algorithm employs four heuristic operators -- expansion, contraction, shrinkage, and reflection -- to explore the search space. Upon meeting predetermined convergence criteria, typically delineated by a convergence sphere, the algorithm transitions into the exploitation phase. Here, it leverages the knowledge obtained from the exploration phase to exploit the optimal configuration previously identified.

Data streams present unique characteristics characterized by their infinite flow and non-stationary distribution. Furthermore, these streams are susceptible to concept drift, whereby the underlying data distribution undergoes significant changes over time. Upon detection of such drift events by specialized detectors like ADWIN or DDM, the SPT algorithm restarts the exploration phase, enabling it to adapt to the evolving data dynamics effectively.

The Micro Evolutionary Self Hyper-Parameter Tuning (MESSPT) algorithm, as described by \cite{moya2023improving}, represents a micro evolutionary approach characterized by the utilization of two distinct operational modalities. Primarily, the exploration mode entails the application of two different operators, namely mutation and crossover, facilitating the comprehensive exploration of the search space. Upon satisfying a predefined convergence criterion, typically delineated by a convergence sphere, the algorithm seamlessly transitions into an exploitation mode. Within this phase, emphasis is placed on leveraging the optimal configurations previously identified during the exploration phase. Moreover, in response to the detection of a concept drift, as signalled by detectors such as ADWIN or DDM, MESSPT restarts the optimization process like SPT algorithm.

\subsection{Boundary Constraint Strategies}

Regarding the Boundary Constraint Strategies, we selected the less computationally expensive techniques, like boundary, reflection, centroid, random or wrapper.
In all these five strategies, there are some standard variables: $RV$ signifies the reconstructed value, $CV$ denotes the computed value within the optimisation process, and $Min_{hp}$ and $Max_{hp}$ denote the lower and upper bounds, respectively, of the hyper-parameter $hp$.

In \textbf{Boundary} method, termed as Projection, the assignment of a value to a variable is adjusted to adhere to the specified boundary conditions \cite{zhang2004handling} (see Figure \ref{fig:boundaryx}).

\begin{equation}
RV =
\begin{cases}
CV & \text{if } Min_{hp} \leq CV \leq Max_{hp} \\
Min_{hp} & \text{if } CV \leq Min_{hp} \\
Max_{hp} & \text{if } CV \geq Max_{hp} \\
\end{cases}
\end{equation}

In the \textbf{Reflection} method, the variables that violate boundary constraints are reflected back from the bound by the amount of violations \cite{robinson2004particle} (see Figure \ref{fig:reflectionx}).

\begin{equation}
RV =
\begin{cases}
CV & \text{if } Min_{hp} \leq CV \leq Max_{hp} \\
Min_{hp} + (Min_{hp}-CV) & \text{if } CV \leq Min_{hp} \\
Max_{hp} - (CV-Max_{hp})& \text{if } CV \geq Max_{hp} \\
\end{cases}
\end{equation}

The \textbf{Centroid} method is employed to rectify exterior boundaries, wherein the centroid of an area derived from $k$ previous configurations is computed \cite{juarez2017improved} (see Figure \ref{fig:centroidx}).

\begin{equation} 
RV = \frac{\sum_1^k RVp_{hp}}{k} 
\end{equation}

Where $RVp_{hp}$ symbolizes the rectification value for a particular historic period. The summation is performed over all $k$ historical periods, and the resulting value is divided by $k$ to obtain the average rectification value.

The \textbf{Random} method involves substituting variables lying outside the designated boundary with a randomly generated value falling within the specified boundaries, as outlined by \cite{price2006differential} (see Figure \ref{fig:randomx}). The following mathematical expression governs this process:
\begin{equation} 
RV = Min_{hp} + Unif * (Max_{hp}-Min_{hp}) 
\end{equation}
Where $Unif$ signifies the function that yields a real-valued outcome distributed uniformly within the interval [0, 1].

In the \textbf{Wrapper} methodology, the search space undergoes a wrapping procedure across each dimension. This involves positing that the search space for each variable exhibits a periodic structure, thereby rendering it amenable to treatment as a cyclic domain \cite{zhang2004handling} (see Figure \ref{fig:wrapperx}). Consequently, values exceeding the upper bound of the search space are repositioned within the space defined by the lower bound, as prescribed by the following formula:

\begin{equation}
RV =
\begin{cases}
CV & \text{if } Min_{hp} \leq CV \leq Max_{hp} \\
Max_{hp} - (Min_{hp}-CV) & \text{if } CV \leq Min_{hp} \\
Min_{hp} + (CV-Max_{hp})& \text{if } CV \geq Max_{hp} \\
\end{cases}
\end{equation}

\section{Results}
\label{sec4}

This section describes the integration of algorithms within the RiverML framework \cite{montiel2021river}. The key goal of the experimental setup is to verify the performance implications of the five different boundary constraint methods on two online optimization algorithms. To this end, we integrate the SPT algorithm introduced by \cite{Veloso0MV21} and MESSPT algorithm introduced by \cite{Moya23} into our experimental framework.

\subsection{Evaluation Protocol}
The experimental methodology follows the Prequential evaluation protocol \cite{gama2009}, a common approach in analysing data streams. This protocol describes the sequential presentation of new data instances, initially designated for model testing and subsequently employed for training. Given the absence of static data, this protocol assumes significance in assessing model efficacy, contrasting with the conventional train/test offline paradigm. The resultant findings and subsequent statistical analyses are organized concerning each task, and set of datasets. Table \ref{tab:datasets} provides details about the datasets utilized in the experiments, including the number of instances, features, and the nature of the dataset (real or synthetic).

\begin{table}[!tb]
    \centering
    \scriptsize
    \begin{tabular}{lcccc}
    \toprule
    \multicolumn{5}{l}{\textbf{TASK: Classification}}\\
    \midrule
    Dataset & Type & Drift & Nr. Instances & Nr. Features \\
    \midrule
    ENRON\tnote{1} & Real & - & 1702 & 1000\\ 
    NOMAO\tnote{1} & Real & - & 34465 & 120\\ 
    RandomRBF\tnote{2}  & Synthetic & - & 20000 & 4 \\ 
    Agrawal\tnote{2}    & Synthetic & - & 20000 & 9 \\ 
    Hyperplane\tnote{2} & Synthetic & Yes & 20000 & 2\\ 
    SEA Drift\tnote{2}        & Synthetic & Yes & 20000 & 3\\ 
    \midrule
    \multicolumn{5}{l}{\textbf{TASK: Regression}}\\
    \midrule
    Dataset & Type & Drift & Nr. Instances & Nr. Features \\
    \midrule
    Tetuan\tnote{3} & Real & - & 52417 & 6\\ 
    Metro\tnote{4} & Real & - & 48204 & 9\\ 
    2DPlanes\tnote{2}  & Synthetic & - & 20000 & 10 \\ 
    MV\tnote{2}    & Synthetic & - & 20000 & 10\\ 
    Friedman Drift\tnote{2} & Synthetic & Yes & 20000 & 10\\ 
    Friedman\tnote{2}        & Synthetic & - & 20000 & 10\\ 
    \midrule
    \end{tabular}
    \caption{Datasets}
    \label{tab:datasets}
\end{table}

A preprocessing pipeline methodology featuring adaptive standard scaling for numerical attributes, as well as one-hot encoding for categorical attributes, is initially applied to the data streams aimed at classification or regression tasks. After this preprocessing, Hoeffding Tree-based models, specifically a Classifier and Regressor proposed by \cite{hulten2001mining}, are employed for predictive tasks. Optimisation procedures targeting three hyperparameters of the Hoeffding Tree models, delta (range: (0.00001, 0.0001)), grace period (range: (100, 500)), and tau (range: (0.01, 0.09)), are conducted. Regarding evaluation criteria, classification tasks are assessed using accuracy, while regression tasks are evaluated through the Root Mean Square Error (RMSE).

\subsection{Heuristic-based and Evolutionary-based Optimizers}

In this subsection, we elucidate and analyse the outcomes derived from our experimental setup for the heuristic-based and evolutionary-based optimiser. 

Table \ref{tab:accuhb} illustrates the performance outcomes of various bounding constraint strategies across diverse learning tasks, showcasing the average accuracy or RMSE per dataset using a heuristic-based optimiser.
\begin{table}[!t]
    \centering
    \scriptsize
    \begin{tabular}{lrrrrr}
    \toprule
    \multicolumn{6}{l}{\textbf{TASK: Classification -- Evaluation Metric: Accuracy (Rank)}}\\
    \midrule
    Dataset & Boundary & Reflection & Wrapper & Random & Centroid \\
    \midrule
    ENRON & 2 & 2 & 1 & 2 & 2 \\
    NOMAO & 5 & 4 & 2 & 3 & 1 \\
    RandomRBF & 2 & 3 & 4 & 5 & 1 \\
    Agrawal & 1 & 2 & 1 & 1 & 3 \\
    Hyperplane & 3 & 2 & 1 & 2 & 2 \\
    SEA\_Drift & 5 & 4 & 3 & 1 & 2 \\
    \midrule
    Avg Rank & 3.00 & 2.83 & 2.33 & 2.33 & \textbf{1.83} \\
    \midrule
    \multicolumn{6}{l}{\textbf{TASK: Regression -- Evaluation Metric: RMSE (Rank)}}\\
    \midrule
    Dataset & Boundary & Reflection & Wrapper & Random & Centroid \\
    \midrule
    Tetuan & 3 & 1 & 2 & 4 & 5 \\
    Metro & 1 & 3 & 4 & 5 & 2 \\
    Friedman & 2 & 1 & 3 & 5 & 4 \\
    MV & 4 & 1 & 2 & 3 & 5 \\
    Friedman Drift & 1 & 2 & 2 & 3 & 4 \\
    Planes2D & 1 & 1 & 2 & 3 & 1 \\
    \midrule
    Avg Rank & 2.00 & \textbf{1.50} & 2.50 & 3.83 & 3.50 \\
    \bottomrule
    \end{tabular}
    \caption{SPT: Average Performance (Best scores highlighted with \textbf{bold})}
    \label{tab:accuhb}
\end{table}
It becomes evident that the SPT optimizer exhibits superior performance  when employing distinct boundary constraint strategies for different tasks. Notably, the centroid strategy consistently outperforms others in classification tasks, while the reflection strategy demonstrates superior efficacy in regression tasks. 
Table \ref{tab:accueb} showcases the performance metrics of different bounding constraint strategies across a two different learning tasks, delineating the average accuracy or RMSE for each dataset using evolutionary-based optimizers.
\begin{table}[!t]
    \centering
    \scriptsize
    \begin{tabular}{lrrrrr}
    \toprule
    \multicolumn{6}{l}{\textbf{TASK: Classification -- Evaluation Metric: Accuracy (Rank)}}\\
    \midrule
    Dataset & Boundary & Reflection & Wrapper & Random & Centroid \\
    \midrule
    ENRON & 1 & 1 & 1 & 1 & 1 \\
    NOMAO & 1 & 1 & 2 & 4 & 3 \\
    RandomRBF & 1 & 1 & 1 & 3 & 2 \\
    Agrawal & 1 & 5 & 4 & 2 & 3 \\
    Hyperplane & 1 & 1 & 1 & 1 & 1 \\
    SEA\_Drift & 1 & 3 & 3 & 3 & 2 \\
    \midrule
    Avg Rank & \textbf{1.00} & 2.00 & 2.00 & 2.50 & 2.00 \\
    \midrule
    \multicolumn{6}{l}{\textbf{TASK: Regression -- Evaluation Metric: RMSE (Rank)}}\\
    \midrule
    Dataset & Boundary & Reflection & Wrapper & Random & Centroid \\
    \midrule
    Tetuan & 2 & 2 & 3 & 1 & 4 \\
    Metro & 4 & 5 & 2 & 3 & 1 \\
    Friedman & 2 & 5 & 1 & 3 & 4 \\
    MV & 2 & 1 & 2 & 3 & 4 \\
    Friedman Drift & 4 & 5 & 2 & 3 & 1 \\
    Planes2D & 1 & 1 & 1 & 1 & 2 \\
    \midrule
    Avg Rank & 2.50 & 3.16 & \textbf{1.83} & 2.33 & 2.66 \\
    \bottomrule
    \end{tabular}
    \caption{MESSPT: Average Performance (Best scores highlighted with \textbf{bold})}
    \label{tab:accueb}
\end{table}
Analysing the results, it is clear that the MESSPT optimizer demonstrates improved efficiency in addressing Classification tasks when adopting the Boundary strategy. However, the Wrapper strategy consistently demonstrates superior performance in Regression tasks compared to alternative methods. 

\subsection{How does the optimizers perform using different boundary constraints strategies and how they and react to concept drift?}
\label{subsec:433}


\begin{figure*}[!ht]
     \centering
     \begin{subfigure}[t]{0.45\textwidth}
         \centering
         \includegraphics[width=\textwidth]{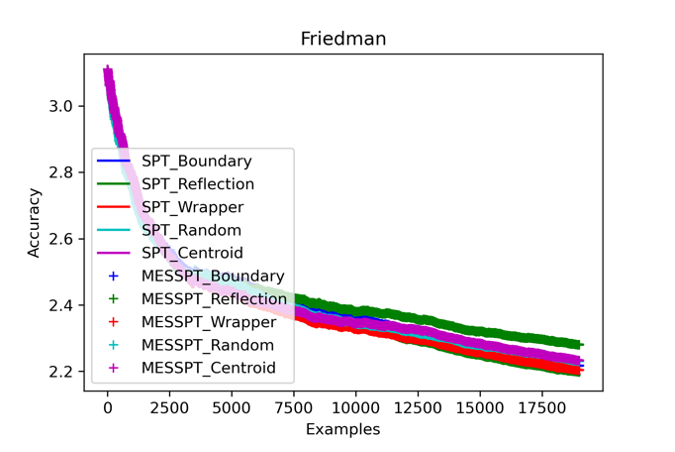}
         \caption{Average Loss - Classification Task}
     \end{subfigure}
     \begin{subfigure}[t]{0.45\textwidth}
         \centering
         \includegraphics[width=\textwidth]{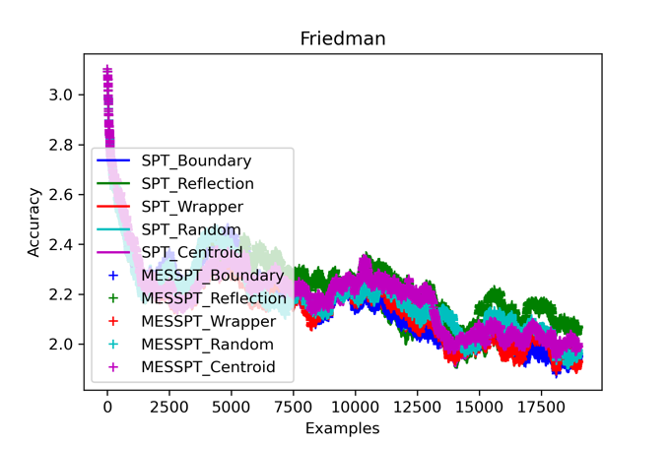}
         \caption{Windowed Loss - Classification Task}
     \end{subfigure}
        \par\bigskip
        \begin{subfigure}[b]{0.45\textwidth}
         \centering
         \includegraphics[width=\textwidth]{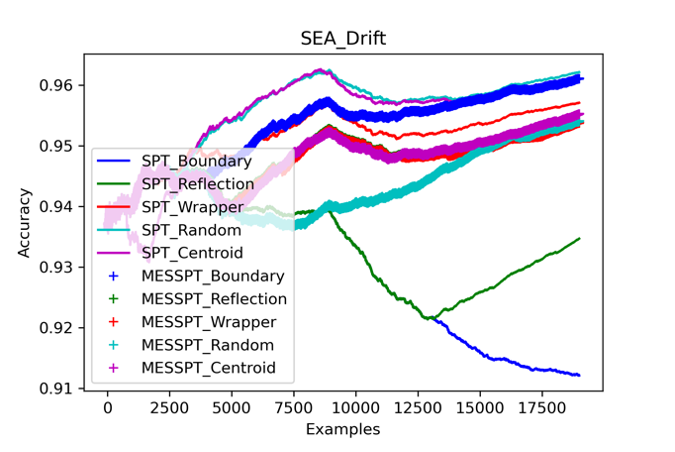}
         \caption{Average Accuracy - Regression Task}
     \end{subfigure}
     \begin{subfigure}[b]{0.45\textwidth}
         \centering
         \includegraphics[width=\textwidth]{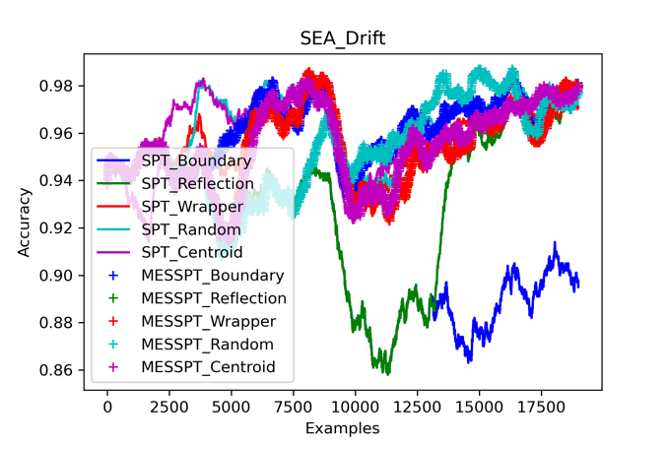}
         \caption{Average Accuracy - Regression Task}
     \end{subfigure}
     \par\bigskip
        \caption{Performance of both optimizers}
        \label{fig:total_scores}
\end{figure*}

We compared both algorithms together (see Figures \ref{fig:total_scores}, 
and the results show that depending on the machine learning task, we can have a superiority of SPT or MESSPT on the top 3 ranked solutions. The MESSPT solution behaves better under drift conditions, also confirmed in the original paper \cite{moya2023improving}.

One of the key features of online learning models is the ability to have a resilient response to concept drift. We present some plots with windowed evaluation metrics for the SEA and Friedman Drift datasets. These datasets hold significant value because they contain abrupt concept drifts and allow us to evaluate the responses of five distinct boundary constraint methods to such dynamics.

Observing Figure \ref{fig:total_scores}, it becomes evident that, in the context of the Classification task, the SPT algorithm demonstrates a substantial performance enhancement in addressing concept drift instances when employing the Random, or Centroid strategies. However, by adopting an evolutionary-based optimiser, optimal strategies for mitigating concept drift behaviour are wrapper and boundary strategies.

Both the SPT and MESSPT methodologies integrate the ADWIN drift detection mechanism to restart the optimisation process. While this design feature may introduce latency in detecting concept drift instances, our findings suggest that the boundary constraint solutions underperform when compared with the other constraint strategies. This makes it imperative for further exploration of constraint strategies to achieve better performance on the evaluation metrics. Subsequently, in the context of the regression task, heuristic-based optimisation algorithms exhibit superior performance, particularly with strategies such as random, boundary, and wrapper. In contrast, evolutionary approaches demonstrate enhanced efficacy with wrapper and centroid strategies.

\section{Conclusions}
\label{sec5}

This study integrated Boundary constraint strategies into two state-of-the-art online optimisation algorithms. These strategies were designed to rectify invalid configurations produced by the optimisers. Specifically, a selection of techniques, including randomisation, centroid adjustment, reflection, and wrapper methodologies, was implemented within both hyper-parameter optimisation algorithms. This study represents a first approach to evaluate the influence of boundary constraint strategies on online hyper-parameter optimisation methods.

Regardless of the optimisation algorithm utilised, our empirical observations shows that the boundary constraint strategies have impact on optimiser efficacy. 
The results shows that the "Reflection" and "Random" strategies does not bring improvements on the performance of the optimizers. The other strategies can add some improvement but highly depends on the optimizer used and machine learning task to solve. Such factors may be related to the optimiser's inherent characteristics or the efficacy of the boundary constraint strategy itself.

These findings creates new research possibilities, emphasising the need to explore more sophisticated boundary constraint strategies for enhancing online optimisation algorithms.
%